%% file: main.tex
\documentclass{article}
\usepackage{spconf,amsmath,epsfig}
\usepackage{amssymb}
\usepackage{upgreek}
\usepackage{amsfonts}
\usepackage{makecell}
\usepackage{array}
\usepackage{units}
\usepackage{multirow}
\usepackage{booktabs}
\usepackage{makecell}
\usepackage{tabularx}
\usepackage{url}
\usepackage[square,numbers]{natbib}

\title{Weakly Supervised Semantic Segmentation of Remote Sensing Images for Tree Species Classification based on Explanation Methods}
\name{Steve~Ahlswede,~Nimisha~Thekke~Madam,~Christian~Schulz,~Birgit~Kleinschmit~and~Begüm~Demir}
\address{Technische~Universitaet~Berlin,~Germany}
\begin{document}
%
\maketitle
\begin{abstract}

The collection of a high number of pixel-based labeled training samples for tree species identification is time consuming and costly in operational forestry applications. To address this problem, in this paper we investigate the effectiveness of explanation methods for deep neural networks in performing weakly supervised semantic segmentation using only image-level labels. Specifically, we consider four methods: i) class activation maps (CAM); ii) gradient-based CAM; iii) pixel correlation module; and iv) self-enhancing maps (SEM). We compare these methods with each other using both quantitative and qualitative measures of their segmentation accuracy, as well as their computational requirements. Experimental results obtained on an aerial image archive show that: i) considered explanation techniques are highly relevant for the identification of tree species with weak supervision; and ii) the SEM outperforms the other considered methods. The code for this paper is publicly available at \url{https://git.tu-berlin.de/rsim/rs_wsss}.

\end{abstract}

\input{01_introduction}
\input{02_methods}
\input{03_results}
\input{04_conclusion}
\vspace{-0.1cm}

\section{Acknowledgments}
This work is funded by the European Research Council (ERC) through the ERC-2017-STG BigEarth Project under Grant 759764 and by the German Federal Ministry of Education and Research through TreeSatAI project under Grant 01lS20014A. We also thank the State Forest of Lower Saxony for providing the aerial images and stand level data. 


\bibliographystyle{IEEEbib}
\small
\bibliography{defs, bibliography}

\end{document}

%% file: 01_introduction.tex
\vspace{-0.1cm}
\section{Introduction}
\vspace{-0.2cm}
Accurate identification of tree species by the analysis of remote sensing (RS) images is important for various forestry applications. By reducing cost intensive on-site surveys, it significantly supports public authorities, conservation agencies and private owners in forest mapping and management. Tree species mapping can be achieved by using semantic segmentation methods, which aim to predict pixel-wise classification results on RS images. Deep learning (DL) based semantic segmentation methods have recently seen a rise in popularity in the context of tree species classification \cite{schiefer2020mapping}. Most DL models require a high amount of labeled samples to optimize all parameters and reach a high performance of tree species classification. The labeling of samples can be achieved based on: 1) in situ ground surveys; 2) the expert interpretation of color composites (image photo-interpretation); or 3) hybrid solutions where both photo-interpretation and ground surveys are exploited \cite{demir2011AL}. Collection of a sufficient number of high quality pixel-level labels associated to tree species can be time consuming, complex and costly.

To address this problem, in this paper we focus our attention on weakly supervised semantic segmentation (WSSS) for tree species segmentation, which rely on weak supervision (i.e., image-level labels). The use of image-level labels can significantly reduce the annotation cost and effort in forestry applications. However, such  supervision indicates only the existence of certain tree species assigned to images without their exact pixel based location information (which is essential for obtaining tree species maps). To obtain accurate tree species maps at the pixel-level, in this paper we investigate the effectiveness of explainable neural networks in the context of WSSS. Explanation methods are capable of generating explanations in the form of pixel-level heatmaps that can be highly relevant for providing a pixel-level tree species map from a DL model trained using image level labels. In this paper, we consider four explanation methods: i) class activation maps (CAM) \cite{zhang2018adversarial}; ii) Gradient-based CAM (GradCAM) \cite{selvaraju2017grad}; iii) pixel correlation module (PCM) \cite{wang2020self}; and iv) self-enhancing maps (SEM) \cite{zhang2020rethinking}. All considered methods have been experimentally compared in terms of their: 1) capability to perform accurate tree species mapping; 2) model complexity; and 3) segmentation time. This work is the first, to the best of our knowledge, that explores the potential of pixel-level explanation heatmaps for tree species mapping in a weak supervision framework.

\vspace{-0.2cm}

%% file: 02_methods.tex
\vspace{-0.1cm}
\section{Tree Species Mapping from Image Level Labels}\label{proposed_sect}
\vspace{-0.2cm}
Let $ \mathcal{X}=\{\textbf{X}_1,\cdots,\textbf{X}_M\} $ denote an archive of $ M $ RS images acquired over forestry areas, where the $ m $-th image within the archive is represented by $ \textbf{X}_m $. We assume that a training set $\mathcal{X}_T \subset \mathcal{X}$ is available where each image $\textbf{X}_n$ within the set is associated with a multi-label vector $\textbf{y}_n = [y_{n, 1}, \cdots, y_{n, S}]$, where $S$ is the total number of classes, indicating which tree species are present. Here, $y_{n, s}$ takes the value of 1 if the class $s$ is present in $\textbf{X}_n$, and 0 otherwise. The set of all multi-label vectors corresponding to $\mathcal{X}_T$ is thus denoted as $\mathcal{Y}_T$.  

To obtain tree-species maps by using image-level labels (and thus weak supervision), we consider a convolutional neural network (CNN) which learns a mapping $ \hat{\textbf{y}}_n=U(\textbf{X}_n)$, where $U(\cdot)$ represents the CNN. However, the predicted output vector $ \hat{\textbf{y}}_n $ by the direct application of a CNN does not provide any spatial information regarding class locations or extents, which are required for tree species mapping. To obtain the map at pixel-level, we investigate CAM, GradCAM, PCM, and SEM in the context of WSSS and exploit the class specific heatmaps (which show the probability of a given class at each pixel) derived by these methods. Specifically, feature maps $\textbf{F} \in \mathbb{R}^{C \times H \times W}$ ($C$, $H$ and $W$ represent the channels, height, and width of the feature maps, respectively) from the final layer of the trained model $U(\cdot)$ corresponding to an image $\textbf{X}_n$ are exploited. The importance of each feature map with respect to each class is derived and a weighted linear combination of the feature maps form the class specific heatmaps. Each class specific heatmap has its values normalized between 0-1, thus acting as pixel probabilities. Heatmaps are thus used to obtain the semantic segmentation by assigning class labels to pixels where the class has the highest probability. Specific details on how the heatmaps are derived from the four methods are given in the following sections.



\subsection{Class Activation Maps}

CAMs aim to highlight the image regions which are used by the CNN to classify an image to a given class. This is achieved by establishing a relationship between the set of feature maps $\textbf{F}$ and a given class. To this end, a 1x1 convolution layer can be applied after $\textbf{F}$, which takes $\textbf{F}$ as input and outputs $S$ feature maps, generating a set of CAMs ($\textbf{A} \in \mathbb{R}^{S \times H \times W}$) \cite{zhang2018adversarial}. Thus, for a given class $s$, we obtain its CAM as such:
\begin{equation}
    \textbf{A}_s = \sum_{c=1}^{C} w^c_s \textbf{F}^c
    \label{cam}
\end{equation}
where $w^c_s$ is the weight of importance for the $c^{th}$ feature map with respect to the $s^{th}$ class.

The CAM is then averaged and used as the final class prediction scores (image level prediction):
\begin{equation}
    \hat{y}_{n, s} = \frac{1}{HW}  \sum_{i=1}^{H}\sum_{j=1}^{W} \textbf{A}_{s,i,j}
    \label{classscore}
\end{equation}
where $i$ and $j$ are the row and column indices of $\textbf{A}_s$, respectively.

Given this formulation, CAMs can be obtained in an end-to-end manner without any post-processing steps. 

\subsection{Gradient Based CAM (GradCAM)}

GradCAM \cite{selvaraju2017grad} is a reformulation of CAM. Instead of using learned weights from a 1x1 convolution to determine the importance of each feature map, GradCAM uses the average gradient associated with each feature map in $\textbf{F}$. Specifically, we first feed an image $\textbf{X}_n$ through the network in order to obtain the vector of predicted class scores $\hat{\textbf{y}}_n$. Then, we perform back-propagation with respect to a single class prediction $\hat{y}_{n, s}$ in order to obtain the gradients for the feature maps $\textbf{F}$ with respect to $\hat{y}_{n, s}$. Finally, we take the mean of the gradients in order to obtain the importance of feature map $\textbf{F}^c$ for class $s$:
\begin{equation}
    w^c_s = \frac{1}{HW} \sum_{i=0}^{W}\sum_{j=0}^{H}  \frac{\delta \hat{y}_{n,s}}{\delta \textbf{F}^c_{i, j}}
    \label{gcam}
\end{equation}

We can thus obtain the GradCAM for a given class by replacing $w^c_s$ in (\ref{cam}) with the definition given in (\ref{gcam}). One distinct advantage of GradCAM is that it can be applied to any network architecture as it gets the importance weights of each feature map using gradient scores. However, the WSSS cannot be obtained in an end-to-end manner, as it requires the back-propagation post-processing.

We would like to emphasize that both CAM and GradCAM have been recently applied for land cover mapping in RS. For details, we refer the reader to \cite{li2020accurate, chan2021comprehensive}.

\subsection{Pixel Correlation Module (PCM)}

The PCM \cite{wang2020self} aims to enhance CAM in order to improve the segmentation. PCM avoids the need for post-processing as the module is integrated into the network itself, thus allowing for end-to-end processing. The module consists of three parallel 1x1 convolutional layers which takes feature maps from different levels of $U(\cdot)$, along with the original image, as input in order to embed them into latent space. The embedded outputs are then concatenated and passed through one additional 1x1 convolution which gives us a matrix of feature maps $\textbf{K} \in \mathbb{R}^{C\times H \times W}$ which correspond to an image $\textbf{X}_n$ and contain a combination of high and low level features. From this matrix, attention scores ($attn_{q, r}$) are obtained which provide the similarity between the features at two spatial locations (e.g. $q$ and $r$) within $\textbf{K}$ through a self-attention mechanism as:
\begin{equation}
    attn_{q, r} = \frac{\textbf{K}_q^T \textbf{K}_r}{||\textbf{K}_q|| \cdot ||\textbf{K}_r||}
    \label{pcm_score}
\end{equation}
where $q$ and $r$ each represent a pixel location (e.g. $q = (0,0)$, $r = (0,1)$). 

For a given class $s$, the PCM value at pixel location $q$ ($\textbf{P}_{s,q}$) is obtained by taking the weighted sum of all values in $\textbf{A}_s$, where the weight values are the cosine similarity estimated between the feature vector at $q$ and all other feature vectors across the spatial dimension, as in (\ref{pcm_score}). This is be formulated as:
\begin{equation}
    \textbf{P}_{s,q} = \sum_{\forall r} attn_{q, r} \times \textbf{A}_{s,r}
    \label{pcm}
\end{equation}

As opposed to \cite{wang2020self} where PCM is trained via supervision from a Siamese network and equivariant cross regularization loss, our work trains the module as a separate branch. To this end, we placed PCM after the feature extraction backbone and trained it using class predictions made from the output of the module. Thus, when training the model which includes PCM, we use two binary cross entropy loss terms, one from the classification head, and one from the PCM module.
\vspace{-0.1cm}

\subsection{Self-Enhancement Maps (SEM)}
SEM \cite{zhang2020rethinking} also aims to enhance the CAM. SEM works on the principle that feature vectors located within target object regions have higher similarity than those within different class regions. Here, input image $\textbf{X}_n$ is passed through the network $U(\cdot)$ in order to obtain $\textbf{A}$ and the feature maps $\textbf{F}$. Given a class specific $\textbf{A}_s$, $E$ seed coordinates corresponding to locations with the largest values are chosen. The feature vectors corresponding to the locations of the $E$ seed points are then extracted from $\textbf{F}$ along the channel dimension. The cosine similarity defined in (\ref{pcm_score}) is then calculated between the $E$ seed feature vectors and all other feature vectors in $\textbf{F}$. This results in $E$ similarity maps, where the value for a given pixel $r$ within the $e^{th}$ similarity map is the cosine similarity score (\ref{pcm_score}) between feature at $r$ and the $e^{th}$ feature. The final output for the given class is then obtained by taking the maximum value at each pixel across the $E$ similarity maps. 

\input{figures/results_fig}

For each of the methods from the previous four sections, we obtain the final tree species map by applying the argmax function across the outputs of the respective method for classes predicted above a threshold $\tau$ at the image level.

\vspace{-0.1cm}

%% file: figures/results_fig.tex
\begin{figure*}[ht]
    \newcommand{\figwidth}{0.13\linewidth}
    \newcommand{\figheight}{0.95in}
    
    \fboxsep=0pt
    \fboxrule=1pt

    \begin{minipage}[l]{\figwidth}
        \centering
        \centerline{\textbf{ }}\medskip
    \end{minipage}
    \begin{minipage}[c]{\figwidth}
        \centering
        \centerline{\textbf{Image}}\medskip
    \end{minipage}
    \begin{minipage}[c]{\figwidth}
        \centering
        \centerline{\textbf{Reference Map}}
        \medskip
    \end{minipage}
    \begin{minipage}[l]{\figwidth}
        \centering
        \centerline{\textbf{CAM \cite{zhang2018adversarial}}}\medskip
    \end{minipage}
    \begin{minipage}[l]{\figwidth}
        \centering
        \centerline{\textbf{GradCAM \cite{selvaraju2017grad}}}\medskip
    \end{minipage}
    \begin{minipage}[l]{\figwidth}
        \centering
        \centerline{\textbf{PCM \cite{wang2020self}}}\medskip
    \end{minipage}
    \begin{minipage}[l]{\figwidth}
        \centering
        \centerline{\textbf{SEM \cite{zhang2020rethinking}}}\medskip
    \end{minipage}

    \begin{minipage}[c]{\figwidth}
        \centering
        \centerline{\textbf{a)}}\medskip
    \end{minipage}
    \begin{minipage}[c]{\figwidth}
        \centering
        {\centerline{\includegraphics[height=\figheight]{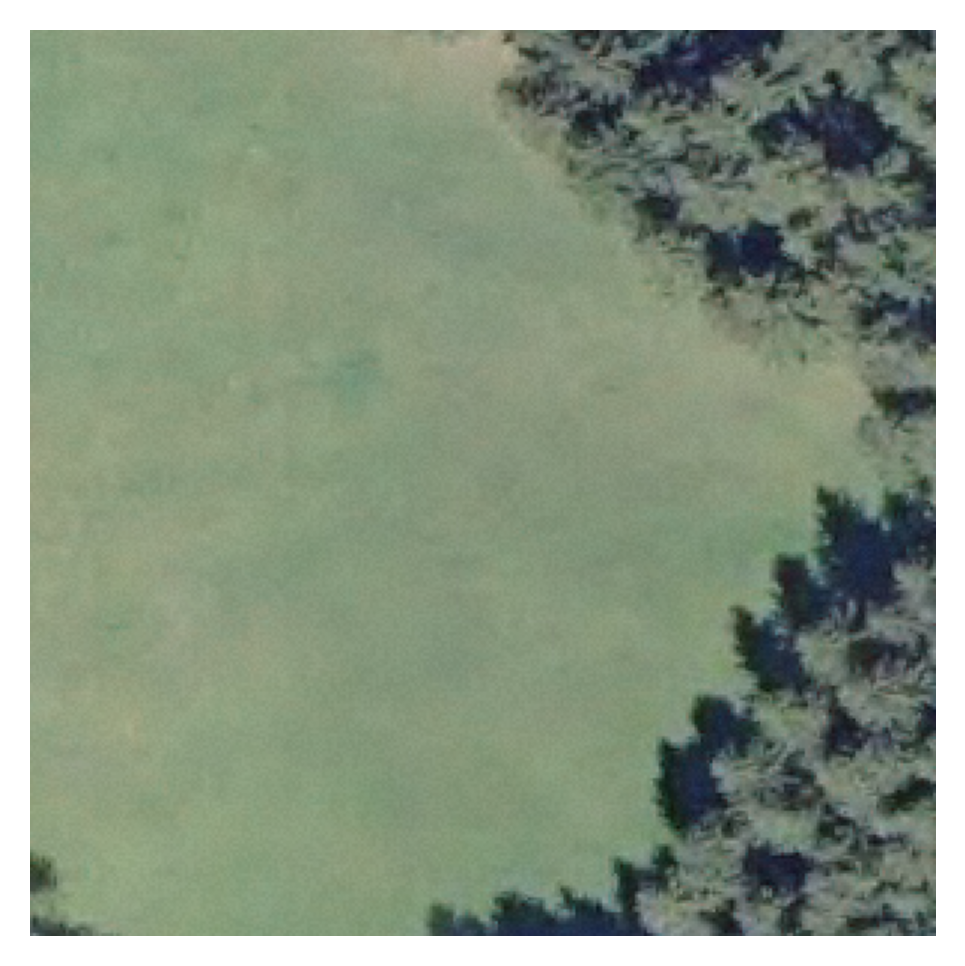}}}\medskip
    \end{minipage}
    \begin{minipage}[c]{\figwidth}
        \centering
        {\centerline{\includegraphics[height=\figheight]{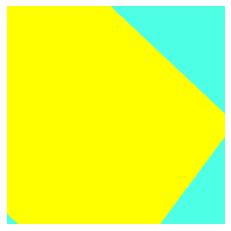}}}\medskip
    \end{minipage}
    \begin{minipage}[c]{\figwidth}
        \centering
        {\centerline{\includegraphics[height=\figheight]{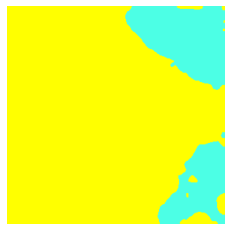}}}\medskip
    \end{minipage}
    \begin{minipage}[c]{\figwidth}
        \centering
        {\centerline{\includegraphics[height=\figheight]{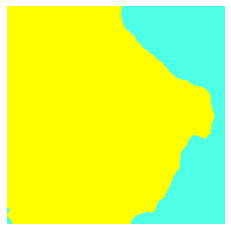}}}\medskip
    \end{minipage}
    \begin{minipage}[c]{\figwidth}
        \centering
        {\centerline{\includegraphics[height=\figheight]{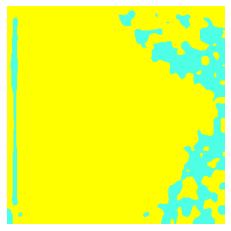}}}\medskip
    \end{minipage}
    \begin{minipage}[c]{\figwidth}
        \centering
        {\centerline{\includegraphics[height=\figheight]{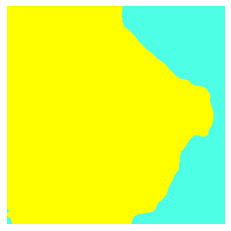}}}\medskip
    \end{minipage}
    
    \begin{minipage}[c]{\figwidth}
        \centering
        \centerline
        \textbf{b)}
        \medskip
    \end{minipage}
    \begin{minipage}[c]{\figwidth}
        \centering
        {\centerline{\includegraphics[height=\figheight]{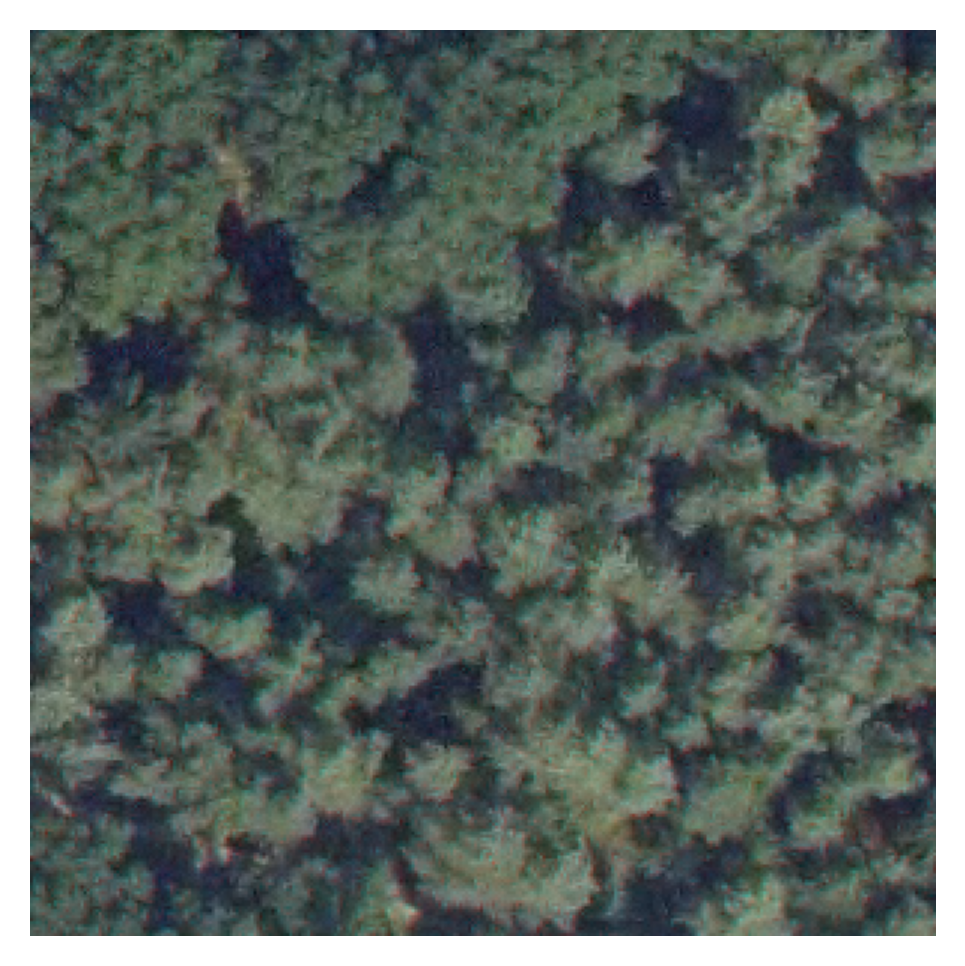}}}\medskip
    \end{minipage}
    \begin{minipage}[c]{\figwidth}
        \centering
        {\centerline{\includegraphics[height=\figheight]{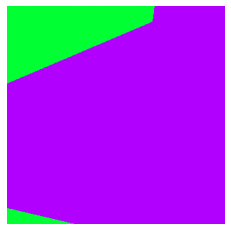}}}\medskip
    \end{minipage}
    \begin{minipage}[c]{\figwidth}
        \centering
        {\centerline{\includegraphics[height=\figheight]{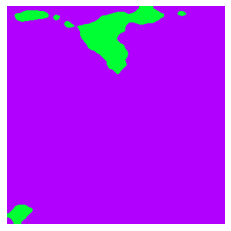}}}\medskip
    \end{minipage}
    \begin{minipage}[c]{\figwidth}
        \centering
        {\centerline{\includegraphics[height=\figheight]{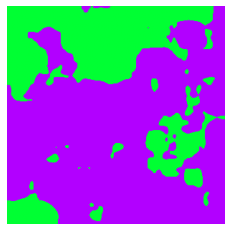}}}\medskip
    \end{minipage}
    \begin{minipage}[c]{\figwidth}
        \centering
        {\centerline{\includegraphics[height=\figheight]{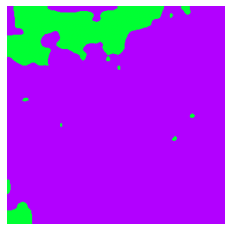}}}\medskip
    \end{minipage}
    \begin{minipage}[c]{\figwidth}
        \centering
        {\centerline{\includegraphics[height=\figheight]{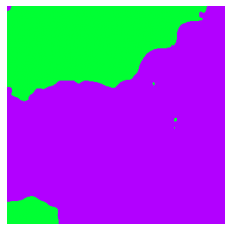}}}\medskip
    \end{minipage}
    
    \begin{minipage}[c]{\figwidth}
        \centering
        \centerline{\textbf{c)}}\medskip
    \end{minipage}
    \begin{minipage}[c]{\figwidth}
        \centering
        {\centerline{\includegraphics[height=\figheight]{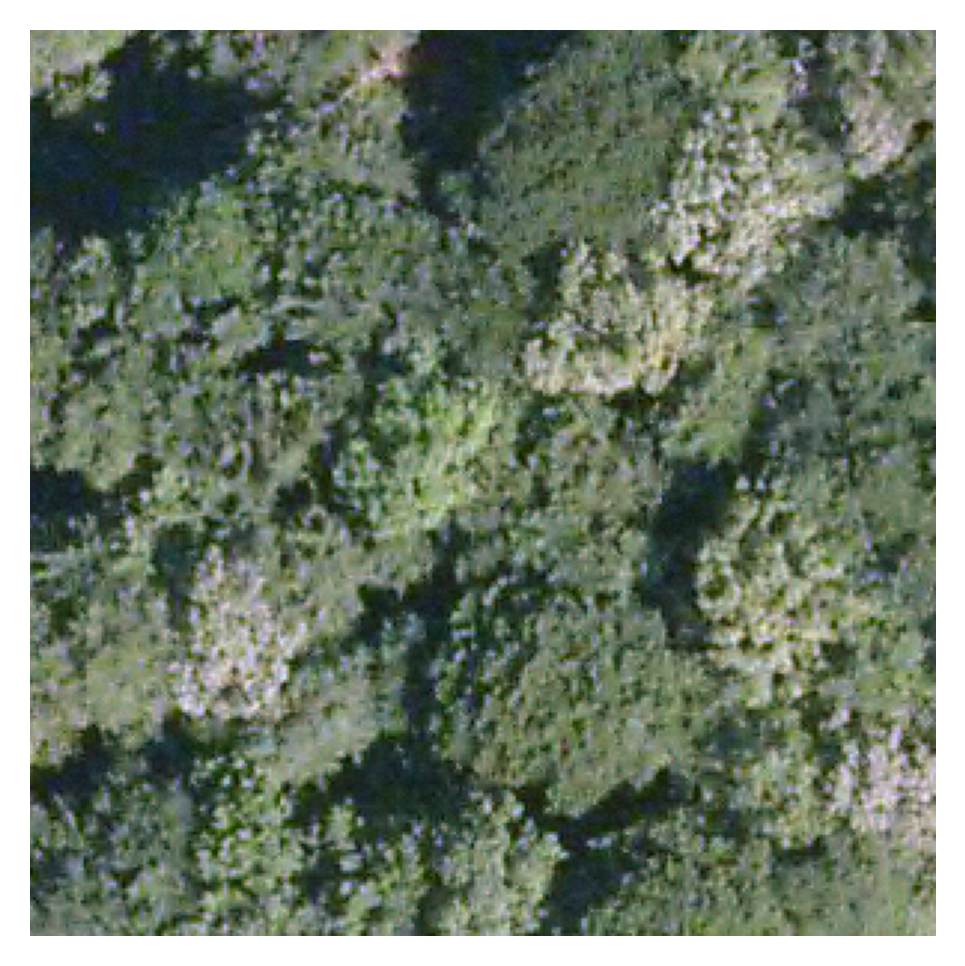}}}\medskip
    \end{minipage}
    \begin{minipage}[c]{\figwidth}
        \centering
        {\centerline{\includegraphics[height=\figheight]{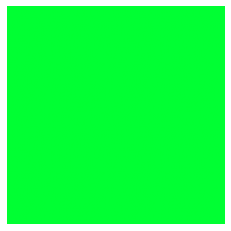}}}\medskip
    \end{minipage}
    \begin{minipage}[c]{\figwidth}
        \centering
        {\centerline{\includegraphics[height=\figheight]{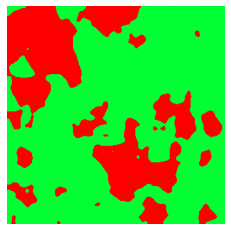}}}\medskip
    \end{minipage}
    \begin{minipage}[c]{\figwidth}
        \centering
        {\centerline{\includegraphics[height=\figheight]{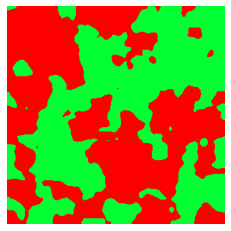}}}\medskip
    \end{minipage}
    \begin{minipage}[c]{\figwidth}
        \centering
        {\centerline{\includegraphics[height=\figheight]{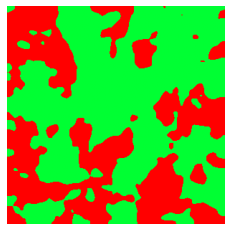}}}\medskip
    \end{minipage}
    \begin{minipage}[c]{\figwidth}
        \centering
        {\centerline{\includegraphics[height=\figheight]{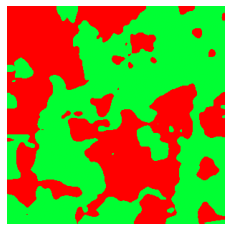}}}\medskip
    \end{minipage}

    \begin{minipage}[c]{\linewidth}
        \centering
        \centerline{\includegraphics[height=0.45in]{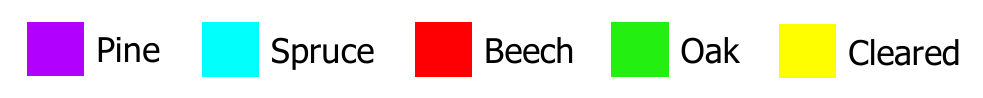}}\medskip
    \end{minipage}
    
    \caption{Examples of images, their reference  and segmentation maps obtained by using different explanation methods in the framework of WSSS.}
    \label{fig:stage1_results}
\end{figure*}

%% file: 03_results.tex
\vspace{-0.1cm}
\section{Experimental Results}
\vspace{-0.2cm}
Experiments were carried out on a dataset acquired from 2012 - 2020 across the German federal state of Lower Saxony. The dataset contains a total of 19,995 aerial images, each of which consists of RGB and near infrared bands with a spatial resolution of 0.2m. The dataset was randomly divided into training (70\%), validation (15\%), and test (15\%) sets. Dominant tree species within stand polygons which overlapped an image patch were used to attain the image-level labels. To evaluate the WSSS results, we rasterized the polygons corresponding to the test set in order to obtain pixel-wise labels. A total of five classes were extracted from the polygons: Pine (\textit{Pinus spp.}), Spruce (\textit{Picea spp.}), Beech (\textit{Fagus spp.}), Oak (\textit{Quercus spp.}), and Cleared. Cleared represents any open areas without tree crown cover (e.g. meadows, clear-cuts, water bodies, etc.).
For the experiments we used DeepLabv3+ \cite{chen2018encoder}. Given that the architecture was designed for semantic segmentation, we replace the segmentation head with a multi-label classification head in order to obtain predictions at the image level.
For evaluation of the WSSS performance we utilized $F_1$, and also examined the number of model parameters (in millions) and the segmentation time per image. All results were obtained using $\tau = 0.5$. The number of seeds used in SEM was chosen empirically based on the performance across the validation image set, resulting in a seed value of $E=10$. 

\input{tables/seg_comp}
From Table \ref{table:comp_wsss} we can see that SEM provides the highest $F_1$ score, whereas CAM leads to the lowest $F_1$ score.
Looking at the qualitative results in Fig. \ref{fig:stage1_results}, we see that CAM is able to correctly localize class regions, but oversegments the dominant class within the image. The performance of CAM is greatly improved when SEM is applied (Fig. \ref{fig:stage1_results}a, b). Predictions made by SEM often follow the contours of tree stands
(Fig. \ref{fig:stage1_results}a, b). 
GradCAM often made predictions visually similar to SEM (Fig. \ref{fig:stage1_results}a). However, GradCAM also predicted incorrect regions (Fig. \ref{fig:stage1_results}b), leading to a degradation in performance.
PCM was able to correctly localize class regions, but tended to oversegment the dominant class within the image, leading to misclassified regions (Fig. \ref{fig:stage1_results}a, b). This was likely due to how the module was trained, as the original implementation involved a Siamese network with a cross-regularization loss. This suggests that PCM may not be a viable addition to standard CNN architectures. 
One area where all methods struggled was when two classes of the same type (e.g. broadleaf or coniferous) were predicted within an image (Fig. \ref{fig:stage1_results}c). Here, the classes are not well separated within the learned feature space of the classifier, leading to misclassifications. 
In regards to segmentation time, CAM had the lowest time, while GradCAM had the highest time. This was due to the need for post-processing step involving a forward and backward pass through the network for each predicted class within an image with GradCAM.
SEM also required a post-processing step, but managed to maintain a low segmentation time. This is because SEM only performs self-attention for a given number of seeds, thus making it less computational than typical self-attention mechanisms. The main drawback with SEM is that more computational time must be invested into tuning the seed hyperparameter, which can become time consuming with larger datasets. With respect to model parameters, only PCM increased model parameters. However, the additional parameters did not achieve any benefits in segmentation performance with respect to GradCAM and SEM.





\vspace{-0.2cm}

%% file: tables/seg_comp.tex
\begin{table}[t]
\vspace{-0.1cm}
\renewcommand{\arraystretch}{0.5}
\centering
\caption{Comparison of the considered explanation methods in terms of $F_1$ scores, the number (in millions) of parameters (\# Param) and the semantic segmentation time (Seg. Time) per image (in millisecond).
}
\label{table:comp_wsss}
\begin{tabular}{lcccc}
\toprule
Metric & CAM\cite{zhang2018adversarial} & GradCAM\cite{selvaraju2017grad} & PCM\cite{wang2020self} & SEM\cite{zhang2020rethinking} \tabularnewline
\toprule
$F_1$ (\%) & 69.88 & 77.18 & 76.70 & \textbf{79.17} \tabularnewline \midrule
\# Param & \textbf{11.65} & \textbf{11.65} & 11.69 & \textbf{11.65} \tabularnewline \midrule
\parbox[l]{12pt}{Seg. Time} & \textbf{50.4} & 402.9 & 54.9 & 54.5 \tabularnewline \midrule
\bottomrule
\end{tabular}
\vspace{-0.4cm}
\end{table}


%% file: 04_conclusion.tex
\vspace{-0.1cm}
\section{Conclusion}
\vspace{-0.2cm}
In this paper, we have studied four explanation methods in the framework of WSSS to obtain pixel-level tree species maps using training samples annotated by image-level labels. In detail, we have investigated: i) CAM; ii) GradCAM; iii) PCM; and iv) SEM based on their semantic segmentation performance, number of model parameters, and semantic segmentation time. The theoretical and experimental analysis show that for tree species segmentation problems, SEM can be chosen as it: i) yields the highest segmentation accuracy; ii) provides the lowest model complexity; and iii) requires a low segmentation time. As a future work, we plan to assess the ability of explanation methods in providing pixel-level pseudo-labels for training a semantic segmentation model.

\vspace{-0.2cm}